\documentclass[conference]{IEEEtran}
\IEEEoverridecommandlockouts
\usepackage{cite}
\usepackage{amsmath,amssymb,amsfonts}
\usepackage{algorithmic}
\usepackage{graphicx}
\usepackage{textcomp}
\usepackage{xcolor}
\usepackage{multirow}
\def\BibTeX{{\rm B\kern-.05em{\sc i\kern-.025em b}\kern-.08em
	T\kern-.1667em\lower.7ex\hbox{E}\kern-.125emX}}
\begin{document}

\title{Reliability of Decision Support \\in Cross-spectral Biometric-enabled Systems
	%Reliability Matrices for Identifying Bias in Facial Recognition
}

\author{\IEEEauthorblockN{ 
		Kenneth Lai$^{1}$, Svetlana N. Yanushkevich$^{1}$, and Vlad Shmerko$^{1}$
	}
	\IEEEauthorblockA{
		\textit{$^1$Biometric Technologies Laboratory, Department of Electrical and Computer Engineering,} \textit{ University of Calgary, Canada,} \\
		Web: http://www.ucalgary.ca/btlab, E-mail: \{kelai,syanshk,vshmerko\}@ucalgary.ca}
	%\newline
	%\textit{$^2$Defence Research and Development Canada (DRDC), Canada.} E-mail: ming.hou@drdc-rddc.gc.ca
}

\maketitle

\thispagestyle{empty}

\begin{abstract}
	This paper addresses the evaluation of the performance of the decision support system that utilizes face and facial expression biometrics. The evaluation criteria include risk of error and related reliability of decision, as well as their contribution to the changes in the perceived operator's trust in the decision. The relevant applications include human behavior monitoring and stress detection in individuals and teams, and in situational awareness system. 
	%Empirical approach was chosen to answer these questions. Main driver of our experimental study is \emph{status protocol} in terms of risk, trust, and bias (R-T-B).
	Using an available database of cross-spectral videos of faces and facial expressions, we conducted a series of experiments that demonstrate the phenomenon of biases in biometrics that affect the evaluated measures of the performance in human-machine systems. 
	
\end{abstract}

%\tableofcontents


%****SECTION*******SECTION*********SECTION*********SECTION****
\section{Introduction}
%****SECTION*******SECTION*********SECTION*********SECTION****

Cross-spectral face images are a valuable source of information for human behavior assessment, in particular, for stress detection of a team \cite{pandey2015distributed}. 

In human-machine systems and teamwork such as a semi-automated military unit, human cognition and autonomous capabilities of machines are combined \cite{johnson2014seven}. The performance of such complex systems are evaluated in different dimensions, e.g. cognitive, technical, psychological, and behavior using notions of Risk, Trust, and Bias (R-T-B) \cite{lai2020performance}. We utilize these and related measures, such as decision reliability, to study biases affecting the performance of face and face expression recognition in cross-spectral bands such as visual, near-infrared (NIR), and infrared (IR). The R-T-B measures a platform for integration of our results that contribute to building trustworthy intelligent systems.

%Underlying principle of our work addressed the common principle of a teamwork : ``If you don’t plan to fail, you fail to
%plan'' \cite{[Johnson-2014]}. In other words, we are interesting the ``side effects'' of cross-spectral facial recognition and facial emotion detection.

%	In Section \ref{sec:Motivation}, we provide motivation, formulate problem and explain the contribution. Related work and basic definitions are in Section \ref{sec:Related-work}. Results of an experimental study are introduced in Section 3. Section 4 concludes the paper.

%****SECTION*******SECTION*********SECTION*********SECTION****
%\section{Motivation, problem, and contribution}
%\label{sec:Motivation}
%****SECTION*******SECTION*********SECTION*********SECTION****

The \textbf{main motivation} of this paper is addressing the looming challenge of building trustworthy intelligent tools, in particular decision support systems. To support the stress assessment mechanisms, additional sources of cross-spectral information can be used.
%Conceptually, the \textbf{problem} is formulated as follows: 
%\emph{Given} a military unit; each team member provides necessary data for assessment performance status including stress state. 
%\emph{The task is} to incorporate facial expression recognition as an additional source of multi-spectral information for stress assessment.
% This task also suggests that the common performance measures such as risk of decision error, trust to decision, and biases affecting the system performance.

The following research questions are the focus of our interest:
\begin{enumerate}
	\item How to assess the risk of decision errors?
	\item How to evaluate the operator's trust in the automated generated decisions? and
	\item How to identify and estimate the potential biases? 
\end{enumerate}
\section{Related work and basic definitions}\label{sec:Related-work}
%****SECTION*******SECTION*********SECTION*********SECTION****

An overview of the studies of cross-spectral face biometrics is provided below.
\begin{small}
	\begin{center}
		%\definecolor{light}{gray}{.9}
		%\colorbox{light}{
		\begin{parbox}[h]{0.95\linewidth} {
				\vspace{-2mm}
				\begin{center}
					\begin{eqnarray}\label{Review}
					\begin{parbox}[h]{0.2\linewidth} {\centering
						\textbf{\texttt{Cross spectral biometric traits}} }
					\end{parbox}
					\equiv\left\{\hspace{-0.2cm}
					\begin{tabular}{ll}					
					\texttt{Face$_{\textit{NIR}}$} & \cite{kang2014nighttime,li2007illumination};\\
					\texttt{Face$_{\textit{IR}}\rightarrow$Face$_{\textit{RGB}}$} &\cite{kezebou2020tr}; \\
					\texttt{Face$_{\textit{RGB}}\leftrightarrow$Face$_{\textit{IR}}$} &\cite{osia2017bridging}; \\														\texttt{Face$_\textit{Bias}$} &\cite{das2018mitigating,orquin2018visual};\\
					\end{tabular} 
					\hspace{-0.2cm}\right \}
					\end{eqnarray}
			\end{center}} 
		\end{parbox}
	\end{center}
\end{small}

The use of near-infrared (NIR) imaging brings a new dimension for face detection and recognition. Motivated by the goal of improving the reliability of face recognition, in \cite{li2007illumination}, an active NIR imaging system has been developed. The NIR illumination was used in \cite{kang2014nighttime} for cross-distance face matching problem in night-time operations. In \cite{kezebou2020tr}, facial images from the IR domain was translated to the RGB domain using GAN. Paper \cite{osia2017bridging} establishes the relationship between facial features in the visual band (RGB image) and infrared (IR) band. In \cite{das2018mitigating,orquin2018visual}, the RGB demographic biases of facial recognition are studied. Note that IR facial image conveys also useful additional information for stress assessment such as arterial pulse \cite{chekmenev2007thermal}.

These papers cover various aspects of cross-spectral face and facial expression recognition. However, none investigate the problem of biases that affect the performance in terms of reliability of decision, risks of error, and their relationship to the trustworthiness of the decision support tool, as perceived by the human operator. This paper represents additional study specific to the cross-spectral biometrics in the context of autonomous and semi-autonomous decision support in human-machine systems. 

The measures of the intelligent tool trustworthiness are related as follows:
\begin{small}
	\begin{equation}
	\begin{array}{c}
	\text{Performance} \\
	\text{Measures} \\
	\end{array}=
	\left\{\begin{array}{l}
	\texttt{Risk}=\text{${F}$(\texttt{Impact}, \texttt{Probability})}\\
	\texttt{Trust} = \mathcal{F}(\texttt{Risk}, \texttt{Reliability})\\
	\texttt{Bias}
	\end{array}
	\right.
	\end{equation}
\end{small}

%\begin{array}{l}
% \textit{Risk} \\
% \textit{Trust} \\
%		\textit{Bias} 
%\end{array}

where \texttt{Probability} represents the error rates of the system, specifically the false non-match rate (FNMR) and the false match rate (FMR). Reliability is similarly defines as True Positive Identification Rate (TPIR). 
%In our study, we use the formalization of Trust as a function of risk and reliability:
%\begin{equation}
%\text{Trust} = \mathcal{F}(\text{Risk},\ \text{Reliability})
%\end{equation}

%Risk is estimated as \text{$\mathcal{F}$(\texttt{Impact}, \texttt{Probability})} 
Risk of error is computed as follows:
\begin{eqnarray} \label{eq:risk}
\texttt{Risk}_\texttt{\textit{Error}}&=& \underbrace{\texttt{Impact}_\texttt{\textit{FNMR}}}_\text{Cost of a FNMR} \times \underbrace{\texttt{Error}_\texttt{\textit{FNMR}}}_{\text{FNMR}} \nonumber\\
&+&\underbrace{\texttt{Impact}_\texttt{\textit{FMR}}}_\text{Cost of a FMR} \times \underbrace{\texttt{Error}_\texttt{\textit{FMR}}}_{\text{FMR}}\\
&=& \alpha \cdot \texttt{Error}_\texttt{\textit{FNMR}} + \beta \cdot \texttt{Error}_\texttt{\textit{FMR}}\nonumber
\end{eqnarray}
where $\alpha$ represents the \texttt{Impact} or cost of a false non-match and $\beta$ represents the \texttt{Impact} or cost of a false match.

In \cite{cohen1998trust}, the definition of trust is dependent on both the qualitative and quantitative aspects of the system. It is defined by the degree of confidence in a system, that is, the ability of the system to produce the correct prediction over many iterations. As such, we can correlate the change in trust with the change in system reliability. While \texttt{Trust} is a function of \texttt{Risk} and \texttt{Reliability}, change in \texttt{Trust}, or \texttt{Bias}, is defined by the difference in decision reliability:
\begin{eqnarray}\label{eq:trust}
\texttt{Bias}_\texttt{\textit{Trust}} &=& \texttt{R}_j - \texttt{R}_i
\end{eqnarray}
where $\texttt{R}_i$ represents the reliability of the system given condition $i$ and $\texttt{R}_j$ represents the reliability of the system given condition $j$. The result can be positive (increase in trust) or negative (decrease in trust).

In this paper, we focus on capturing the change in the degree of \texttt{Trust} which is critical in dynamic human-machine systems, such as combat team-work in ever-changing conditions. As described in \cite{eastwood2018technology}, the performance in a simulated environment is wildly different from real-world applications. To develop an approach to evaluating the trust and its change, we identify the biases and evaluate their influence on decision reliability.

The existence of bias can deeply alter the performance of any classification or identification algorithms. One common bias is attributed to the demographic of the dataset. An example is a dataset containing an imbalanced number of subjects for gender. Given the list of the different cohorts (emotions) and the overall performance of a dataset, we can divide the overall performance into separate performance measures categorized based on the individual cohorts. For example, the overall accuracy of the Tufts Face Database is 86.29\%. This performance can be partitioned based on the cohort, emotions. These partition includes neutral emotion contributing an accuracy of 81.03\%, while the ``wearing sunglasses'' emotion contributes a 98.66\%. In this case, there is a ``bias for'' sunglasses and a ``bias against'' neutral.

\section{Experiments}

In this paper, we developed two sets of experiments to explore the fundamental performance behavior and the relation between performance, trust, risk, and bias of a machine learning approach such as deep neural networks (NN). 

The first experiment is designed to analyze the performance of the NN based classifier for subject-based identification, which is a 1:N comparison to find the true identity within a selected cohort. We measure the performance in terms of identification rate, specifically the True Positive Identification Rate (\texttt{TPIR}), also known as \texttt{Reliability}. It is defined as follows \cite{grother2019face}:
\begin{equation}\label{eq:tpir}
\texttt{TPIR} = 1-\frac{\text{Number of subjects outside top R ranks}}{\text{Number of searches}}
\end{equation}
In general, the NN classifier prediction is based on the highest value obtained from the output of the softmax layer. This prediction method corresponds to the rank-1 prediction, which corresponds to using only the singular and highest score for prediction.

The second experiment is based on evaluating the model's capability in performing emotion classification in both the RGB and infrared spectra. The decision making performance is evaluated using \texttt{Accuracy}, \texttt{Sensitivity}, and \texttt{Specificity} defined as follows:
\begin{equation}\label{eq:acc}
\texttt{Accuracy}=\frac{TP+TN}{TP+FN+TN+FP}
\end{equation}
\begin{equation}\label{eq:sp}
\texttt{Sensitivity}=\frac{TP}{TP+FN}
\end{equation}
\begin{equation}\label{eq:se}
\texttt{Specificity}=\frac{TN}{FP+TN}
\end{equation}
where $TP$ represents true positive (correct prediction of emotion), $FN$ denotes false negative (incorrect prediction of emotion), $TN$ indicates true negative (correct rejection of emotion), and $FP$ represents false positive (incorrect prediction of emotion). 

\subsection{Dataset}
\begin{figure}[!ht]
	\begin{center}
		\begin{tabular}{cc}
			\includegraphics[width=0.13\textwidth]{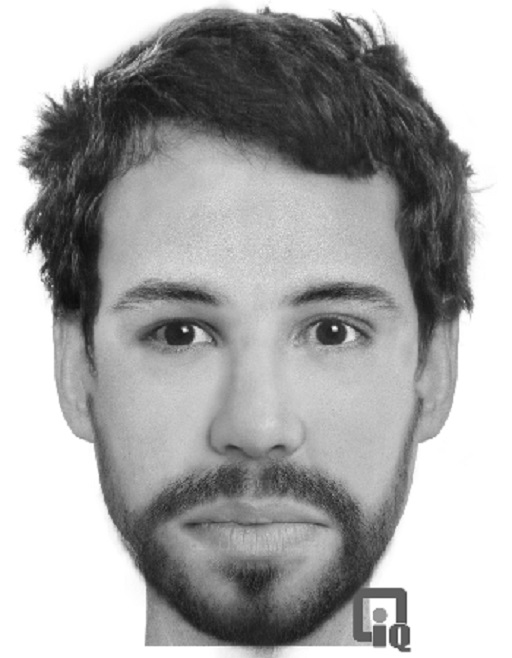} &
			\includegraphics[width=0.22\textwidth]{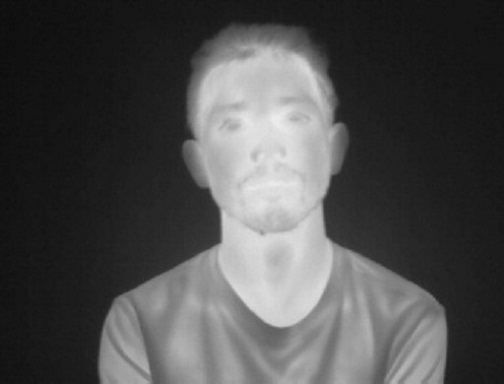} \\
			(a) & (b) \\
			\includegraphics[width=0.22\textwidth]{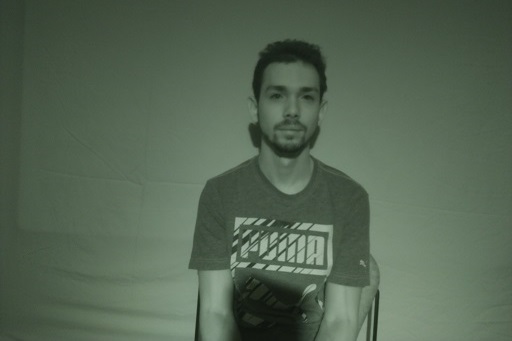} &
			\includegraphics[width=0.22\textwidth]{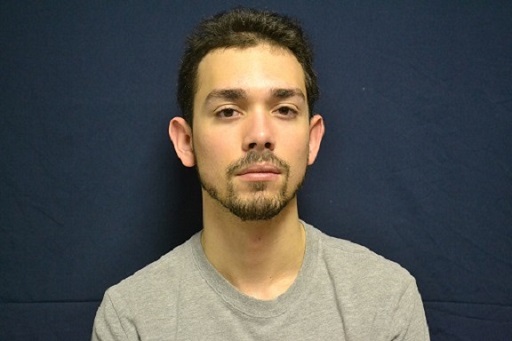} \\
			(c) & (d) \\
		\end{tabular}
		\caption{Sample images of different modalities from the Tufts Face Database \cite{panetta2018comprehensive}: (a) Sketch, (b) Infared, (c) Near-Infared, and (d) RGB.}
		\label{fig:img}
	\end{center}
\end{figure}

The Tufts Face Database \cite{panetta2018comprehensive} contains over 10 000 images from 113 individuals. This database was chosen for evaluation because images of the same subject were collected with different sensors allowing for a better evaluation of cross-modality algorithms such as face recognition. Modalities include thermal, 3D, near-infrared, RGB color, and computerized sketches (shown in Figure \ref{fig:img}).

\begin{figure}[!ht]
	\begin{center}
		\includegraphics[width=0.45\textwidth]{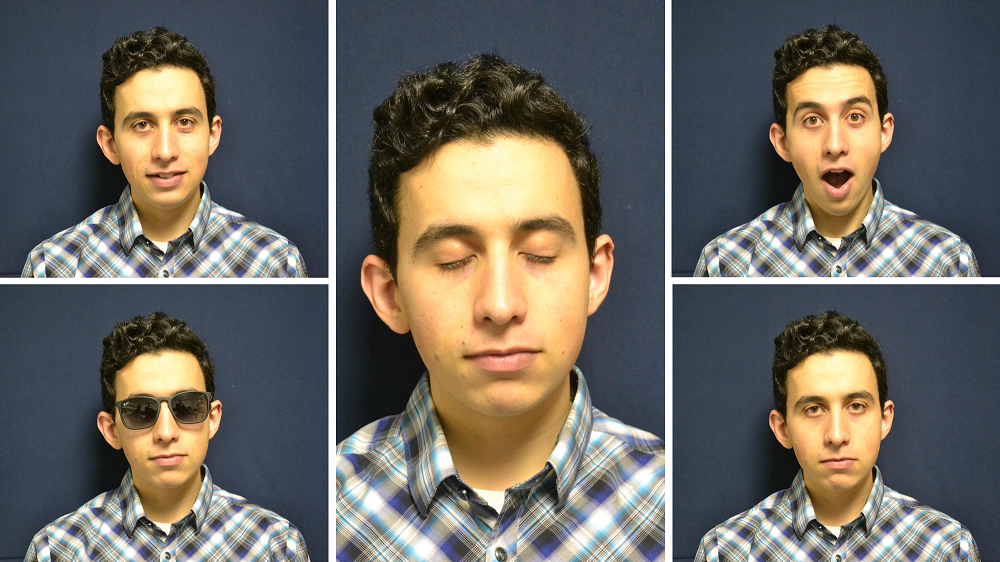} 
		\caption{Sample expression images from the Tufts Face Database \cite{panetta2018comprehensive}. Top left: Smile, Bottom Left: Wearing Sunglasses, Middle: Sleepy, Top Right: Shock, Bottom Right: Neutral.}
		\label{fig:expression}
	\end{center}
\end{figure}

The dataset is separated into different partitions according to the emotion expressed in the image. Performance is evaluated accordingly, dependent on the type of classification (subject vs. emotion). The emotions present in the dataset include neutral, smile, sleepy, shock, and ``wearing sunglasses''. Example images of the 5 expressions are shown in Figure \ref{fig:expression}. 

\subsection{Subject Identification}
For identification, the NN based classifier is first used to extract features. It is based on the ResNet-50 model, pre-trained using the VGGFace dataset \cite{parkhi2015deep}. The extracted features are then used to perform identification. The NN for subject identification is trained using a stochastic gradient descent approach with a learning rate of $0.0001$ and a momentum of $0.9$ for 10 epochs. The performance measures are computed based on an iterative Five-fold-cross-validation, where each fold represents a unique emotion. For each iteration, one fold/emotion is used for testing, while the remaining folds are used for cross-validation.

Table \ref{tab:emotions} shows the classifier performance in terms of \texttt{TPIR} for the Tufts Face Database. The performance is separated into unique partitions depending on the emotion present. The rows in the table represent the partition used for testing, while the columns represent the partition used for validation. The diagonal values in the table are excluded as they represent the special condition where the validation set is the same as the testing set. For example, given row 5 (Shock) and column 4 (Sleepy), the \texttt{TPIR} obtained is 97.32\%. This \texttt{TPIR} represents the scenario of using the Shock partition for testing, Sleepy partition for validation, and the remaining emotions (neutral, smile, sunglasses) for training.
\begin{table}[!htb]
	\centering
	\caption{Reliability (TPIR) values for RGB-Based Identification with Emotion Biases}\label{tab:emotions}
	\begin{tabular}{@{}cc|ccccc@{}}
		&	&	\multicolumn{5}{c}{Validation Set}	\\											
		&	Emotions	&	Neutral	&	Smile	&	Sleepy	&	Shock	&	Sunglasses	\\
		\hline														
		\multirow{5}{*}{\rotatebox[origin=c]{90}{Testing Set}} &	Neutral	&	 - 	 & 	 1.0000 	 & 	 1.0000 	 & 	 1.0000 	 & 	 1.0000 		 \\ 
		&	Smile	&	 0.9911 	 & 	 - 	 & 	 0.9911 	 & 	 0.9732 	 & 	 0.9911 	 	 \\ 
		&	Sleepy	&	 1.0000 	 & 	 0.9911 	 & 	 - 	 & 	 1.0000 	 & 	 0.9911 	 	 \\ 
		&	Shock	&	 0.8929 	 & 	 0.9375 	 & 	 0.9732 	 & 	 - 	 & 	 0.9643 	 	 \\ 
		&	Sunglasses	&	 0.3571 	 & 	 0.3839 	 & 	 0.3482 	 & 	 0.4732 	 & 	 - 	 	 \\ 
		\hline											
		&	Average	&	 0.8103 	 & 	 0.8281 	 & 	 0.8281 	 & 	 0.8616 	 & 	 0.9866 	 \\ 
	\end{tabular}
\end{table}

The \texttt{Reliability} values show that the performance of the identification model is influenced by different emotions. Specifically, the \texttt{TPIR} for row 6 (Sunglasses) is very low which indicated that the identification model is not as capable in identifying subjects ``wearing sunglasses'' when there the training data contains no subjects ``wearing sunglasses''. This observation illustrates a bias embedded in the identification model which is otherwise hidden in the reported averaged \texttt{TPIR} (last row).

Table \ref{tab:modality} reports the \texttt{TPIR} obtained at each rank interval (1, 5, and 10) for different modalities. Modalities of interest in this paper are thermal (IR), near-infrared (NIR), color (RGB), and sketches. The three panels (rank-1, rank-5, rank-10) in Table \ref{tab:modality} represent the \texttt{Reliability} matrices at each rank. Rank-1, 5, and 10 performance are defined as the performance obtained when the top 1, top 5, and top 10 scores are used for prediction, respectively. Each row represents the modality used for training, while each column represents the modality used for testing. For example, in row 2 (RGB) and column 3 (NIR), a TPIR of 93.58\% is obtained when the model is trained with RGB images, tested with NIR images, and evaluated using rank-1 prediction.

\begin{table*}[!htb]
	\centering
	\caption{Reliability (TPIR) for Cross-Modality Subject Identification}\label{tab:modality}
	\begin{tabular}{@{}c|cccc|cccc|cccc@{}}
		
		& \multicolumn{4}{c|}{Rank-1}	&	\multicolumn{4}{c|}{Rank-5}	&	\multicolumn{4}{c}{Rank-10}	\\																						
		&	RGB	&	NIR	&	IR	&	Sketch	&	RGB	&	NIR	&	IR	&	Sketch	&	RGB	&	NIR	&	IR	&	Sketch	\\
		\hline	
		RGB	&	 1.0000 	&	 0.9358 	&	 0.0218 	&	 0.0342 	&	 1.0000 	&	 0.9849 	&	 0.0803 	&	 0.1368 	&	 1.0000 	&	 0.9942 	&	 0.1432 	&	 0.1966 	\\
		NIR	&	 0.7569 	&	 0.9951 	&	 0.0058 	&	 0.0513 	&	 0.9037 	&	 0.9977 	&	 0.0623 	&	 0.2051 	&	 0.9407 	&	 0.9991 	&	 0.1150 	&	 0.3162 	\\
		IR	&	 0.0356 	&	 0.0148 	&	 0.9917 	&	 0.0085 	&	 0.0963 	&	 0.0749 	&	 1.0000 	&	 0.0513 	&	 0.1538 	&	 0.1481 	&	 1.0000 	&	 0.0769 	\\
		Sketch	&	 0.0345 	&	 0.0177 	&	 0.0083 	&	 1.0000 	&	 0.1256 	&	 0.0790 	&	 0.0398 	&	 1.0000 	&	 0.2078 	&	 0.1429 	&	 0.0970 	&	 1.0000 	\\
	\end{tabular}
\end{table*}

Note that in \cite{kezebou2020tr}, a cross-spectral face identification performance of 88.65\% is obtained via synthetically creating RGB images from IR images using a Thermal-to-RGB Generative Adversarial Network (TR-GAN).

There are three observations pertaining to the \texttt{Reliability} of the cross-spectral face identification. The first observation is that the diagonal values of the matrices show the best performance. This describes the idea that using the same modality of data for both training and testing data is beneficial for reaching the highest performance. The second observation is the strong correlation between the RGB and NIR images and a minimal correlation between the other modalities. The last observation is the asymmetry of the RGB and NIR performance, that is, the model trained on RGB and tested on NIR is significantly better (93.58\%) than vice versa (75.69\%). This observation is likely due to: (a) the image similarity between these two spectra of images and (b) the pre-training of the architecture using the VGGFace dataset which contains purely RGB-based images.

\begin{figure*}[!ht]
	\begin{center}
		\begin{tabular}{cc}
			\includegraphics[width=0.45\textwidth,interpolate]{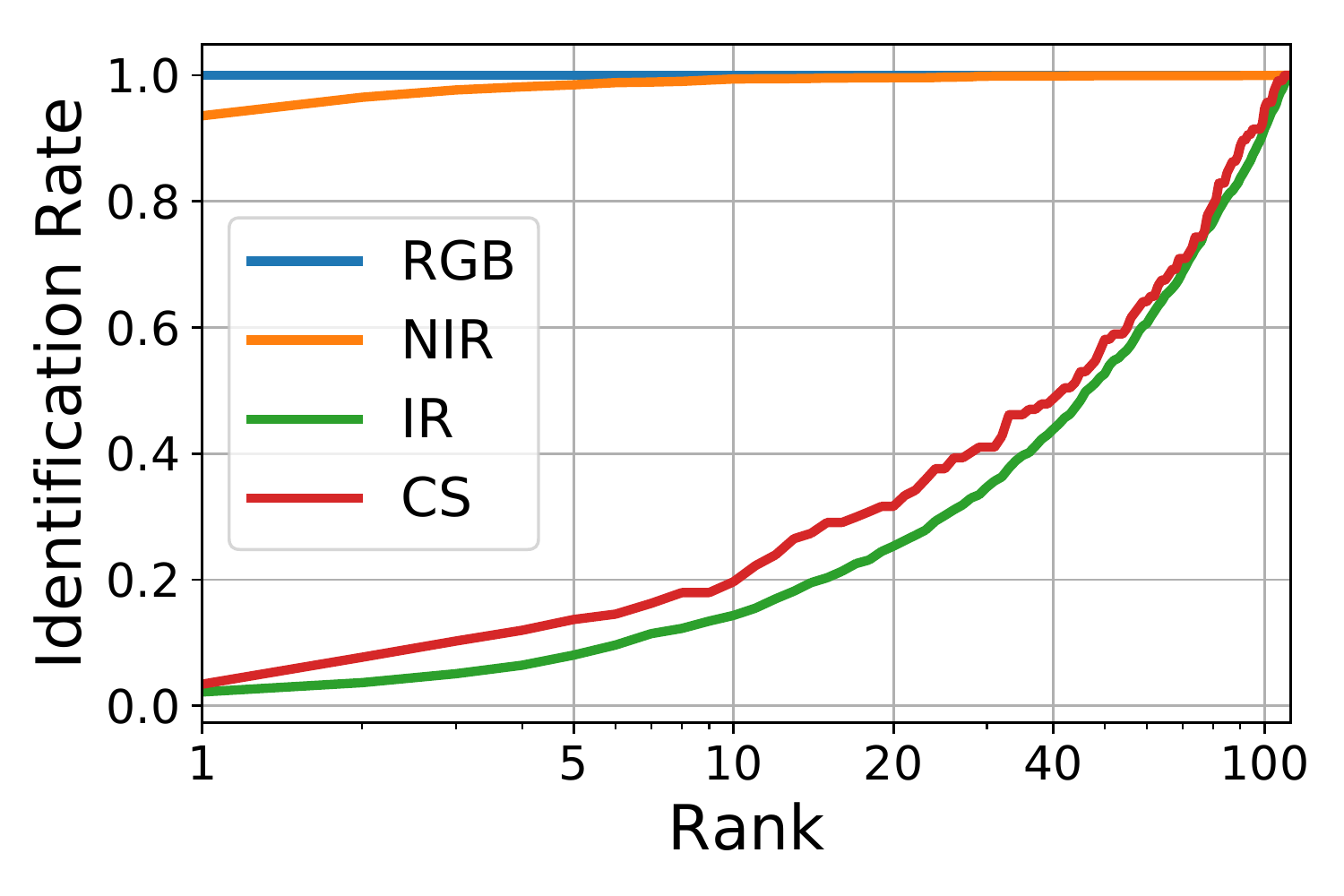} &
			\includegraphics[width=0.45\textwidth,interpolate]{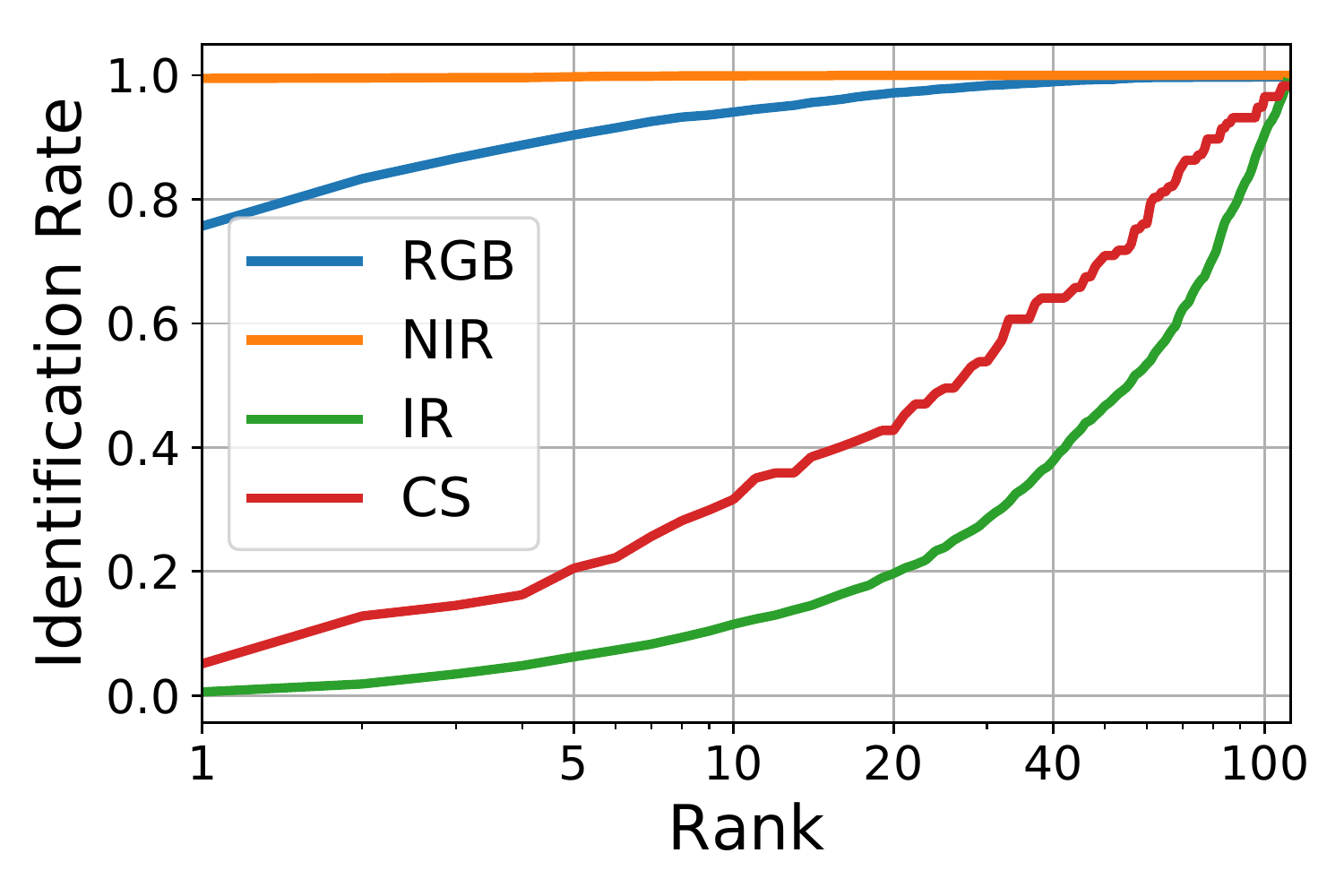} \\
			(a) & (b) \\
			\includegraphics[width=0.45\textwidth,interpolate]{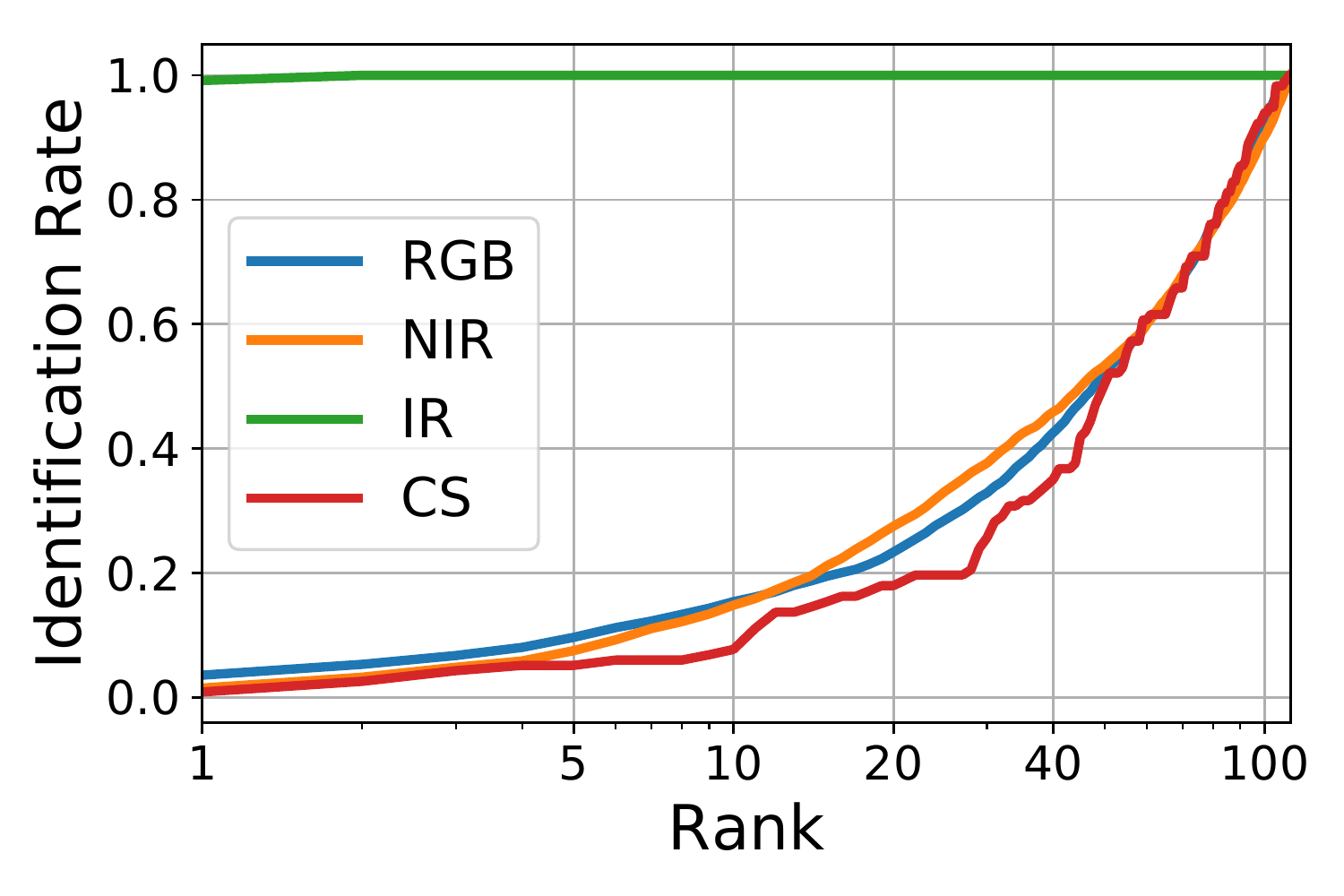} &
			\includegraphics[width=0.45\textwidth,interpolate]{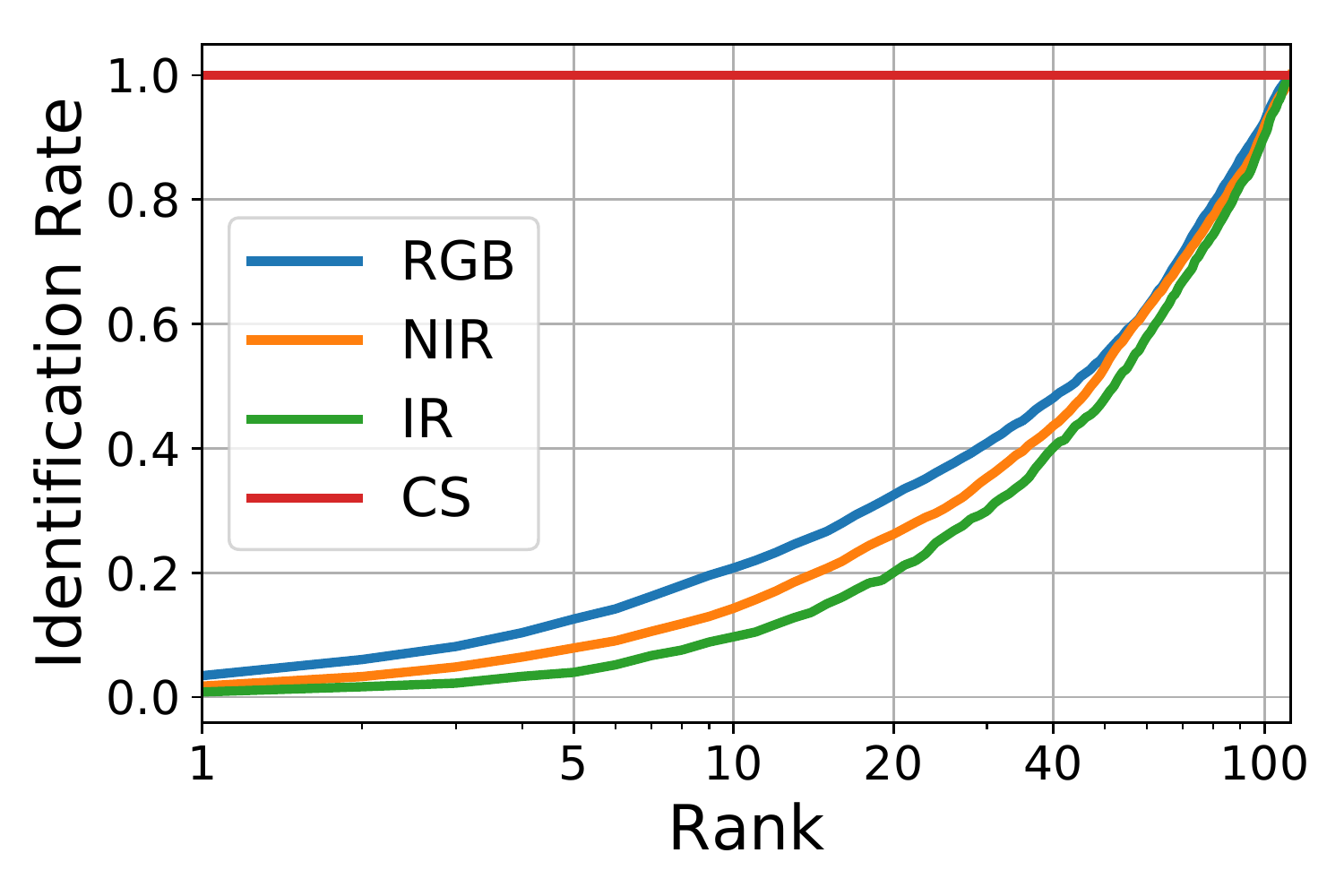} \\
			(c) & (d) \\
		\end{tabular}
		
		\caption{The CMC for subject identification trained only with (a) RGB images, (b) NIR images, (c) IR images, and (d) sketch (CS) image tested using the different modalities (RGB, NIR, IR, and sketch). The blue, orange, green, and red curves represent the performance obtained using RGB, NIR, IR, and sketch images for testing, respectively.}
		\label{fig:cmc}
	\end{center}
\end{figure*}

Figure \ref{fig:cmc} shows the cumulative matching characteristic (CMC) curves on the Tufts Face dataset for subject identification. Figure \ref{fig:cmc}(a) and (b) reveals a linear relationship between the performance of RGB and NIR images, specifically as the performance of identifying subjects using RGB images increases, so does NIR images. 
%In addition, Figure \ref{fig:cmc}(b) shows that the performance of sketch images are much better than random guessing. This behavior reveals that a model trained on NIR images extracts selected features that can be used in sketch images
Figure \ref{fig:cmc}(c) and (d) supports the observation that IR and sketch images are independent. As shown as the performance of identifying subjects using IR or sketch images increases, the identification rate for other modalities remains as a very low constant (which is equivalent to random guessing).

In this section, we evaluated the performance of the system in terms of \texttt{TPIR} for subject identification and examined the influence of bias derived from emotions and cross-spectra on the \texttt{Reliability} of the system decision. Given Equation \ref{eq:trust}, change in \texttt{Trust} is a function of \texttt{Reliability}. Complete \texttt{Trust} can be achieved when both the training and testing information is known; however, for real-world applications, selected information is hidden, such as the ``black box'' aspect in certain face recognition systems. For example, given a system trained with RGB images, a \texttt{Reliability} of 1 is obtained when similar RGB images are used for testing; however, if the desired objective is to use other modalities for testing, such as NIR, then the change in \texttt{Trust}, or $\texttt{Bias}_\texttt{\textit{Trust}}$, is computed as follows:
\begin{eqnarray}
\texttt{Bias}_\texttt{\textit{Trust}} &=& \texttt{R}_\texttt{\textit{NIR}}-\texttt{R}_\texttt{\textit{RGB}}\nonumber\\
&=&0.93358-1.0000=\fbox{-0.06642} \nonumber
\end{eqnarray}
The change in \texttt{Trust} changing RGB images to NIR images is \fbox{-0.06642}, representing an overall net loss of \texttt{Trust}. The change in testing procedures results in a less reliable system. The computation of $\texttt{Bias}_\texttt{\textit{Trust}}$ can be used in combination with the Technology Gap Navigator \cite{eastwood2018technology} to calculate the trust in a system when the base condition is modified, such as the deployment in a different environment.

\subsection{Emotion Classification}
Similar to the subject identification model, a ResNet-50 NN model is also used for emotion classification. The emotion classifier is trained using stochastic gradient descent with a learning rate of $0.0001$ and a momentum of $0.9$ for 100 epochs. Unlike subject identification, the performance measures for emotion classification is computed using a subject-based five-fold-cross-validation, where each fold contains 22-23 subjects.

Table \ref{tab:emotionsc} presents the \texttt{Accuracy}, \texttt{Sensitivity}, and \texttt{Specificity} performances for emotion classification on the Tufts Face Database. Each row indicates the method used for evaluation and each column specifies the performance measure used. In this experiment, there are four variants of the testing procedure: colored-based classification with the four base emotions (RGB:4), colored-based classification with the four base emotions + ``wearing sunglasses'' (RGB:4+S), infrared-based classification with the four base emotions (IR:4), and infrared-based classification with the four base emotions + ``wearing sunglasses'' (IR:4+S). The last row indicates the state-of-the-art classification rate using the Thermal Emotion Recognition System (TERNet) proposed by \cite{km2019ternet}. Note that TERNet operates on pre-processed IR images and is designed to perform classification on the four base emotions. 

\begin{table}[!htb]
	\centering
	\caption{Rank-1 Performance for Emotion Classification}\label{tab:emotionsc}
	\begin{tabular}{@{}c|ccc@{}}										
		&	Accuracy	&	Sensitivity	&	Specificity	\\
		\hline
		RGB:4	&	 0.9420 	 $\pm$ 	 0.1162 	 & 	 0.9420 	 $\pm$ 	 0.0403 	 & 	 0.9803 	 $\pm$ 	 0.0171 	 \\ 
		RGB:4+S	&	 0.9679 	 $\pm$ 	 0.0830 	 & 	 0.9679 	 $\pm$ 	 0.0402 	 & 	 0.9918 	 $\pm$ 	 0.0069 	 \\ 
		IR:4	&	 0.7411 	 $\pm$ 	 0.2395 	 & 	 0.7403 	 $\pm$ 	 0.0999 	 & 	 0.8957 	 $\pm$ 	 0.0099 	 \\ 
		IR:4+S	&	 0.8085 	 $\pm$ 	 0.1769 	 & 	 0.8082 	 $\pm$ 	 0.1477 	 & 	 0.9456 	 $\pm$ 	 0.0324 	 \\ 
		\hline
		TERNet \cite{km2019ternet}&	 0.9620	 & 	 - 	 & 	-	 \\ 
	\end{tabular}
\end{table}

Table \ref{tab:emotionsc} shows that the performance of classifying four emotions is worst than classifying five emotions. In general, the addition of new classes increases the complexity necessary to maintain the same performance. This behavior is not observed in Table \ref{tab:emotionsc}, mainly because the new emotion is very unique. The previous four emotions consist of neutral, smile, sleepy, and shock, which are very similar; however, the new addition ``wearing sunglasses'' provides new and better details, allowing for better fine-tuning of the classification model. For example, the previous model trained on only the 4 emotions focuses details on the eyes, but when the new emotion is added, the emphasis on eye features is reduced due to the sunglasses covering the eyes. This phenomenon showcases a bias in the dataset.

\begin{figure}[!ht]
	\begin{center}
		\begin{tabular}{cc}	\hspace{-3mm}
			\includegraphics[width=0.22\textwidth]{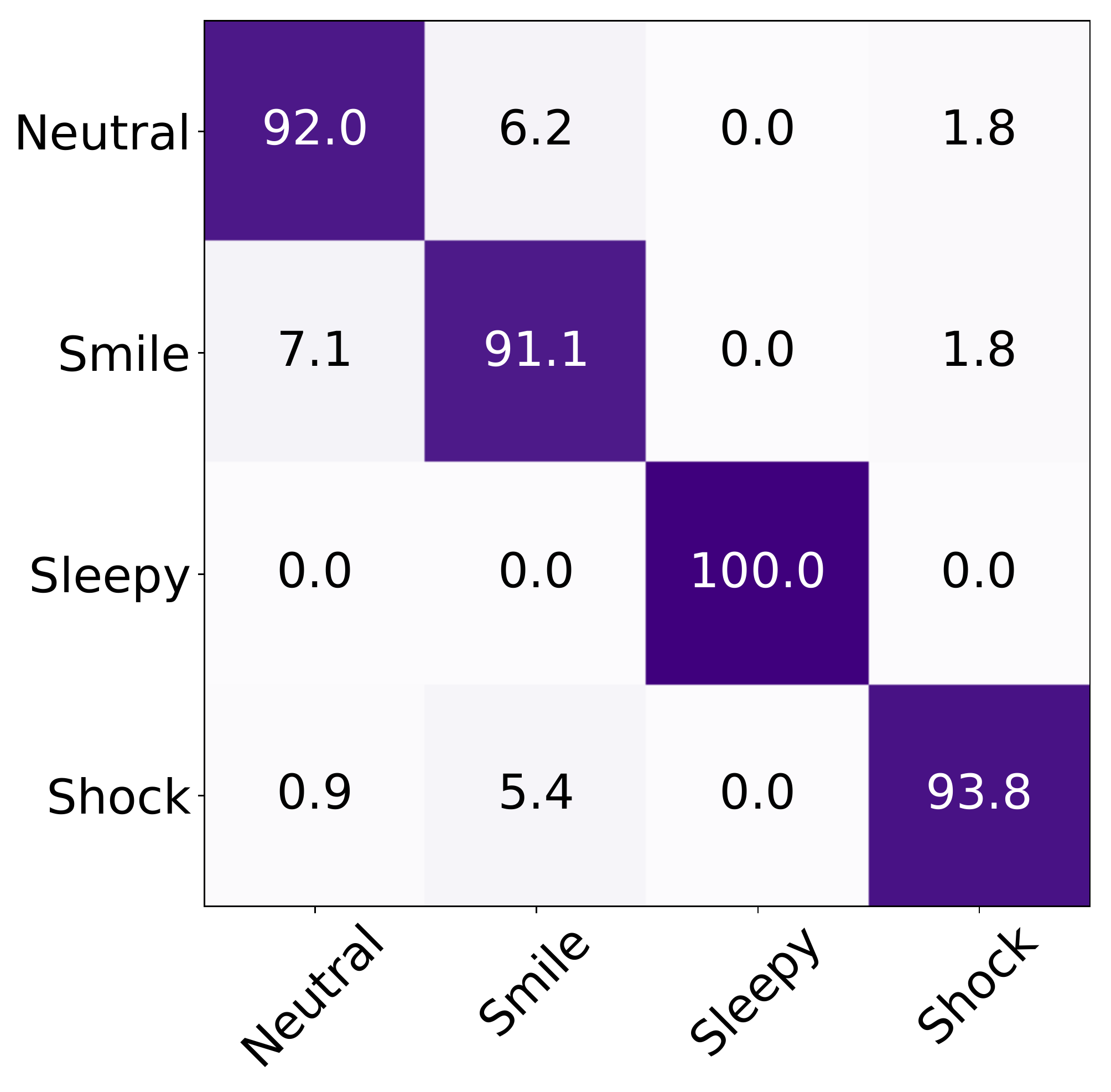} &
			\includegraphics[width=0.24\textwidth]{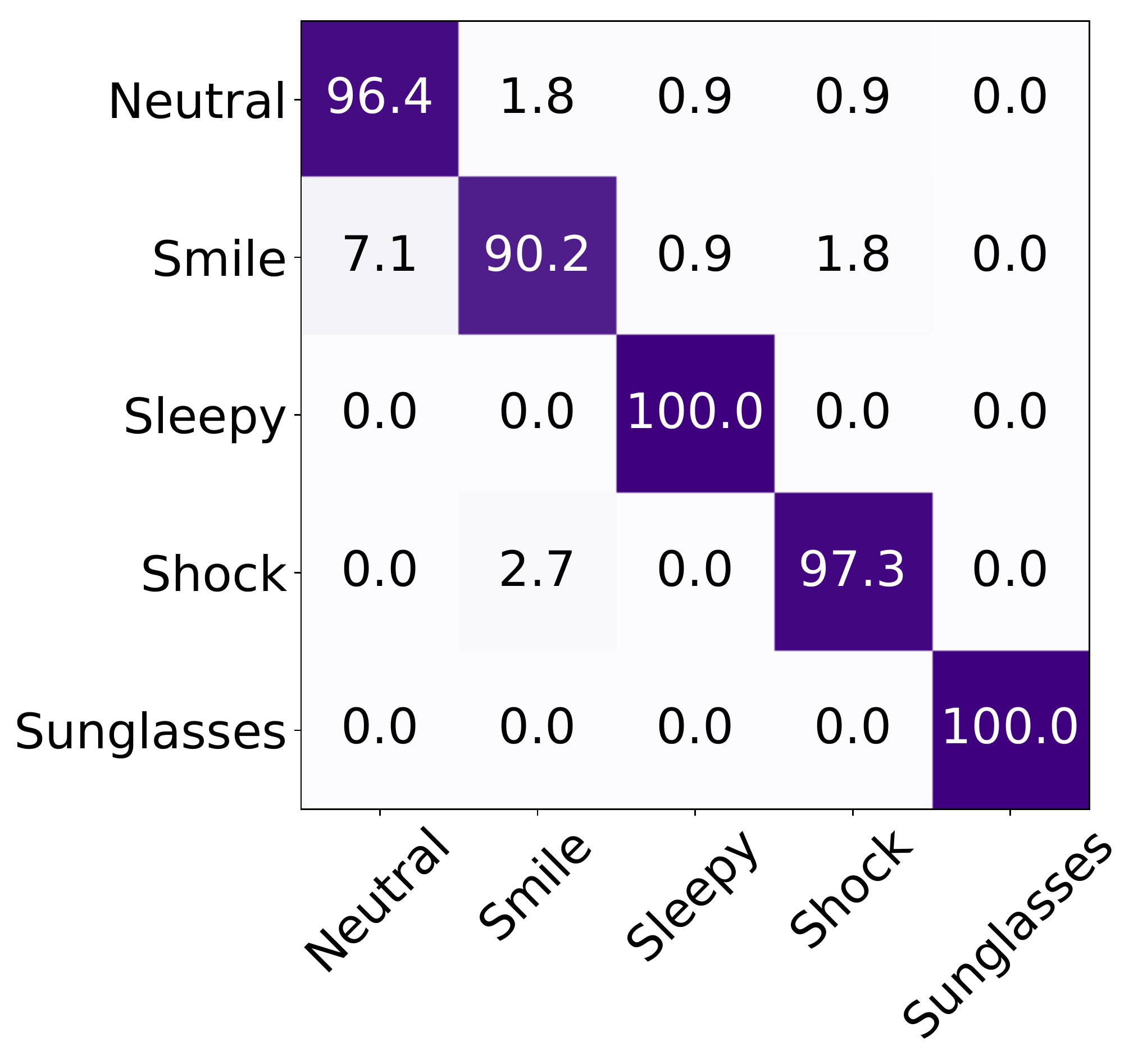} \\
			(a)	&	(b)	\\
		\end{tabular}
	\end{center}
	\caption{Confusion matrix for RGB-based emotion classification: (a) base emotions (neutral, smile, sleepy, shock) and (b) base emotions + ``wearing sunglasses''. The rows indicate the ground-truth emotions, while the columns represent the predicted emotions.
	}\label{fig:rgbcm}
\end{figure}

The confusion matrices shown in Figure \ref{fig:rgbcm} represents the results of emotion classification using only RGB images. Each row in Figure \ref{fig:rgbcm}(a) and (b) represents the ground-truth emotions, and each column indicates the emotion predicted by the classification model. Figure \ref{fig:rgbcm}(a) and (b) contains the accuracy for the 4 base emotion and the four base emotion + ``wearing sunglasses'' emotion, respectively. 

As mentioned previously, a special phenomenon is noticed when comparing the performance of four emotions with 4+S emotions. With the addition of the ``wearing sunglasses'' emotion, every other emotion except smile increases in accuracy. Comparing the performance of smile (row 2) in Figure \ref{fig:rgbcm} (a) and (b), indicates that 0.9\% of smile accuracy has been shifted to the sleepy emotion. A possible explanation is that the images containing sunglasses provide additional data that encourage the classification model to learn better identifiable features, which increases the overall accuracy of the system (from 94.20\% to 96.79\%) at the cost of reducing the accuracy of a specific emotion, smile (from 91.1\% to 90.2\%).

\begin{figure}[!ht]
	\begin{center}
		\begin{tabular}{cc}	\hspace{-3mm}
			\includegraphics[width=0.22\textwidth]{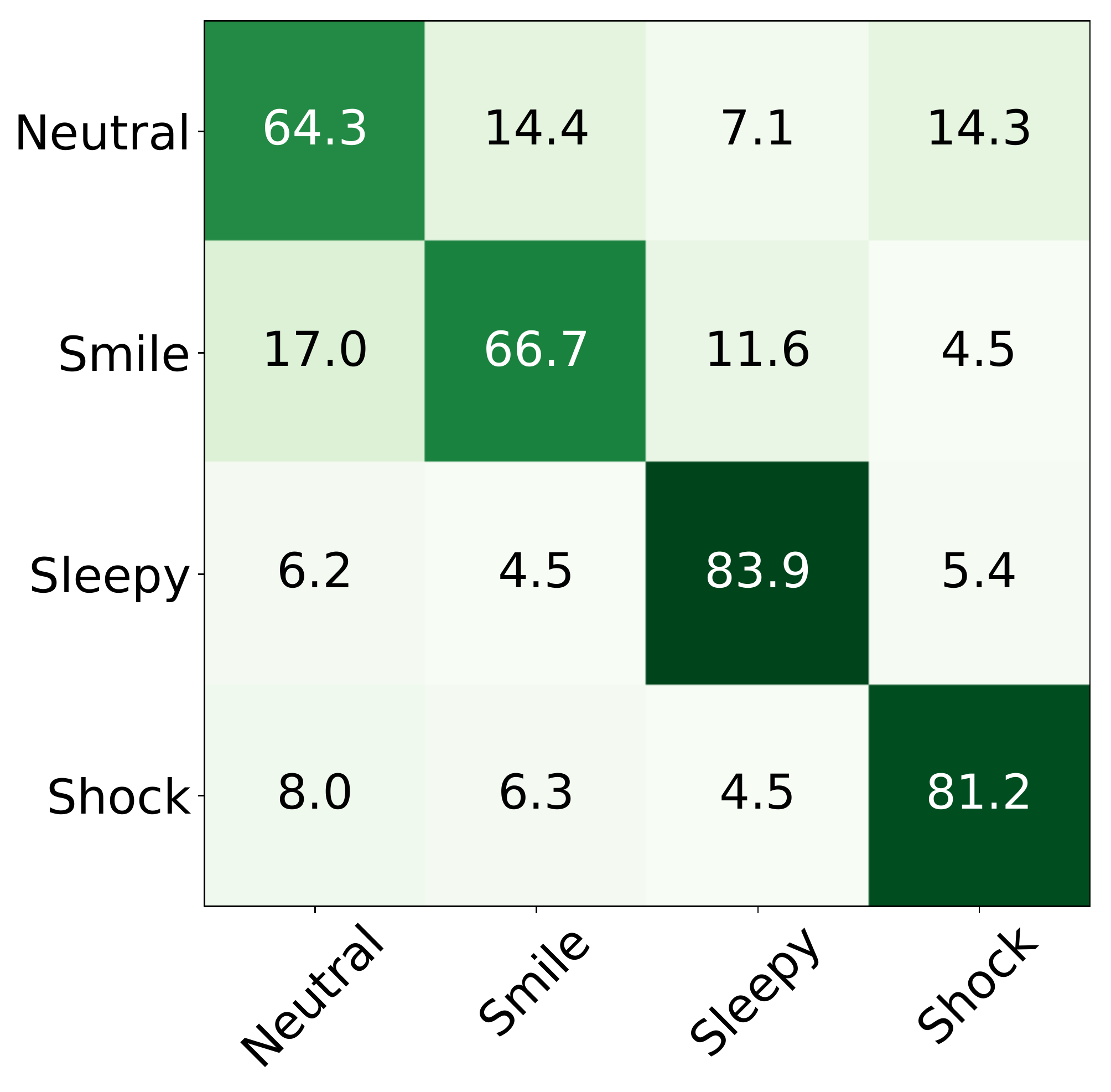} &
			\includegraphics[width=0.24\textwidth]{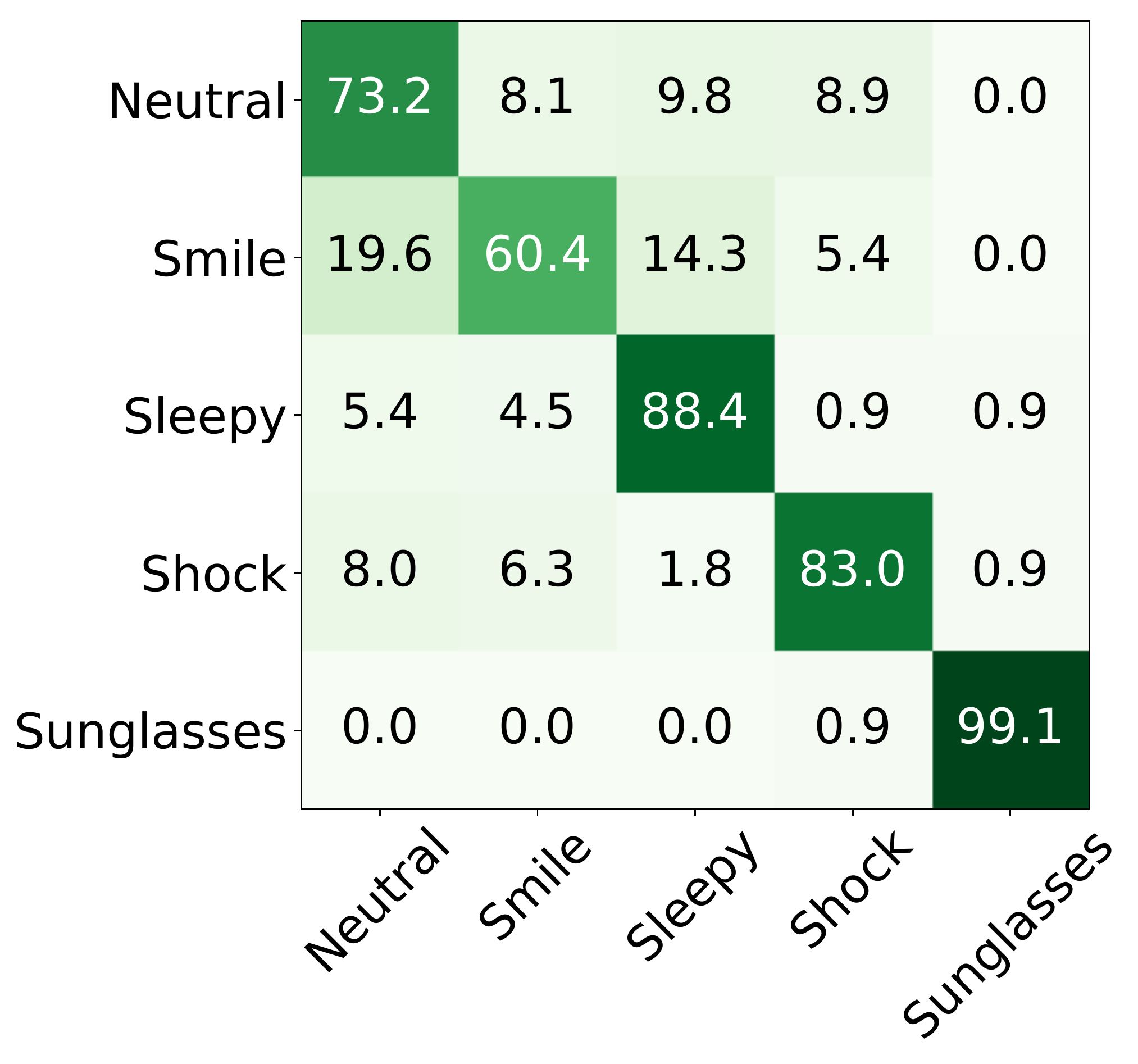} \\
			(a)	&	(b)	\\
		\end{tabular}
	\end{center}
	\caption{Confusion matrix for IR-based emotion classification: (a) base emotions (neutral, smile, sleepy, shock) and (b) base emotions + ``wearing sunglasses''. The rows indicate the ground-truth emotions, while the columns represent the predicted emotions.
	}\label{fig:ircm}
\end{figure}

Similarly, Figure \ref{fig:ircm} shows the confusion matrix of emotion classification using IR images. The same behavior that exists for the RGB images is observed for the IR images, the addition of the ``wearing sunglasses'' emotion increases the accuracy of the other emotions except for smile. A special attribute of IR images is that objects such as glasses and sunglasses appear black in IR images. This unique attribute may be one of the reasons why the classification accuracy of ``wearing sunglasses'' is near perfect (99.1\%). Future works can exploit this behavior to further improve the performance of the system.

In this section, we performed emotion classification on the Tufts Face database. The result of emotion classification reveals a training bias. The classification model is biased towards the four base emotions which can be alleviated by providing new data that are uniquely different in order to prompt the model to learn better features, such as ``wearing sunglasses''.

Alternately, a risk of error can be computed using the \texttt{Sensitivity} and \texttt{Specificity} of the model. Given Equation \ref{eq:risk} and a balanced cost ($\alpha=\beta=1$), the \texttt{Risk} of error for RGB:4+S is calculated as follows:
\begin{eqnarray}
\texttt{Risk}_\texttt{\textit{Error}} &=& \alpha \cdot \overbrace{\texttt{Error}_\texttt{\textit{FNMR}}}^{\text{1-\text{Sensitivity}}}+ \beta \cdot \overbrace{\texttt{Error}_\texttt{\textit{FMR}}}^{\text{1-\text{Specificity}}}\nonumber\\
&=&1-0.9679+1-0.9918=\fbox{0.0403}\nonumber
\end{eqnarray}
Note that $\texttt{Error}_\texttt{\textit{FNMR}}$ is equivalent to $1-\texttt{Sensitivity}$ and $\texttt{Error}_\texttt{\textit{FMR}}$ can be represented as $1-\texttt{Specificity}$. The risk associated with RGB:4+S (0.0403) is lower than RGB:4 (0.0777). The difference in risk value represents the influence of adding the ``wearing sunglasses'' emotion to mitigate the training bias.

\section{Discussion and conclusion}
Our study addresses the problem of bias, risk, and trust in a generic neural network customized for face identification and emotion classification. Bias can be derived from the dataset, such as the demographics or inherent cohorts such as emotions. Risk as a function of error and cost can be estimated using the sensitivity and specificity performance measures. Trust in a system can be modified based on the reliability of the identification module.

For face identification, we examined how bias derived from different emotions impact the overall performance of the identification model. In addition, we explore the capability of cross-modality face identification using the same architecture and report the performance in a series of reliability matrices. Lastly, we show how trust can be gained or lost depending on the Reliability (True Positive Identification Rate) of the system.

For emotion classification, we analyzed how the selection of training data can bias the model toward a specific behavior. With a default classification accuracy of 94.20\%, we can increase the accuracy to 96.79\% by contributing additional unique data that encourages the machine learning model to learn better distinguishable features.

In this paper, we identify the existence of bias within a selected database and illustrate how these biases can be exploited to generate trust in the machine learning model. We propose the creation of a reliability matrices to identify the correlation between the different attributes and characteristics in order to improve the performance of the system.

\section*{Acknowledgments}
\begin{small}
	This Project was partially supported by Natural Sciences and Engineering Research Council of Canada (NSERC) through grant ``Biometric-Enabled Identity Management and Risk Assessment for Smart Cities'', and the Department of National Defence’s Innovation for Defence Excellence and Security (IDEaS) program, Canada. 
\end{small}

{\small
	\bibliographystyle{IEEEtran}
	\bibliography{bias}
}

\end{document}